\documentclass[letterpaper, 10 pt, conference]{ieeeconf}
\IEEEoverridecommandlockouts
\usepackage{cite}
\usepackage{amsmath,amssymb,amsfonts}
\usepackage[ruled, vlined, linesnumbered]{algorithm2e}
\usepackage{graphicx}
\usepackage{textcomp}
\usepackage{xcolor}
\usepackage{siunitx}
\usepackage{booktabs}
\usepackage{amssymb}
\usepackage[ruled, vlined, linesnumbered]{algorithm2e}
\def\BibTeX{{\rm B\kern-.05em{\sc i\kern-.025em b}\kern-.08em
T\kern-.1667em\lower.7ex\hbox{E}\kern-.125emX}}
\usepackage{diagbox}
\usepackage{multirow}

\begin{document}

\title{Embedded Flexible Circumferential Sensing for Real-Time Intraoperative Environmental Perception in Continuum Robots
}
 \author{Peiyu Luo, Shilong Yao, Yuhan Chen, Max Q.-H. Meng\textsuperscript{*}, \IEEEmembership{Fellow, IEEE}
\thanks{This work was supported in part by Shenzhen Key Laboratory of Robotics Perception and Intelligence (ZDSYS20200810171800001), High level of special funds (G03034K003) from Southern University of Science and Technology, Shenzhen, China.}
\thanks{P. Luo, S. Yao, and M. Q.-H. Meng are with Shenzhen Key Laboratory of Robotics Perception and Intelligence and the Department of Electronic and Electrical Engineering, Southern University of Science and Technology, Shenzhen, China.}
\thanks{S. Yao is with the Department of Electrical Engineering, City University of Hong Kong, Hong Kong, China.}
\thanks{Y. Chen is with the Institute of Robotics and Intelligent Systems, Dalian University of Technology, Dalian 116024, China.}
\thanks{$*$Corresponding author: Max Q.-H. Meng. e-mail: max.meng@ieee.org.}
}  
\maketitle

\begin{abstract}
Continuum robots have been widely adopted in robot-assisted minimally invasive surgery (RMIS) because of their compact size and high flexibility. However, their proprioceptive capabilities remain limited, particularly in narrow lumens, where lack of environmental awareness can lead to unintended tissue contact and surgical risks. To address this challenge, this work proposes a flexible annular sensor structure integrated around the vertebral disks of continuum robots. The proposed design enables real-time environmental mapping by estimating the distance between the robotic disks and the surrounding tissue, thereby facilitating safer operation through advanced control strategies. The experiment has proven that its accuracy in obstacle detection can reach 0.19 mm. Fabricated using flexible printed circuit (FPC) technology, the sensor demonstrates a modular and cost-effective design with compact dimensions and low noise interference. Its adaptable parameters allow compatibility with various continuum robot architectures, offering a promising solution for enhancing intraoperative perception and control in surgical robotics.
\end{abstract}

\begin{keywords}
Continuum robots, optoelectronic sensor, circumferential perception, minimally invasive surgery.
\end{keywords}

\section{Introduction}
In minimally invasive surgery, continuum robots leverage their highly integrated compact structures and compliant control characteristics to assist surgeons in performing precise operations within narrow tissue spaces and complex anatomical lumens. The core flexible manipulator overcomes the spatial constraints of traditional rigid instruments through unique structure redundancy, enabling safe navigation through tortuous luminal environments. This bioinspired design not only circumvents operational blind spots inherent in rigid architectures but also significantly enhances intraoperative safety via active compliance control. Recent years have witnessed significant advances in research on applying continuum manipulators to intraluminal interventions, particularly in surgical procedures\cite{troncoso2022continuum}\cite{zhang2022design}\cite{gao2024transendoscopic} and modeling for safe control\cite{zhang2024composite}\cite{chen2024chained}.
\begin{figure}[h]
    \centering
    \includegraphics[width=0.5\textwidth]{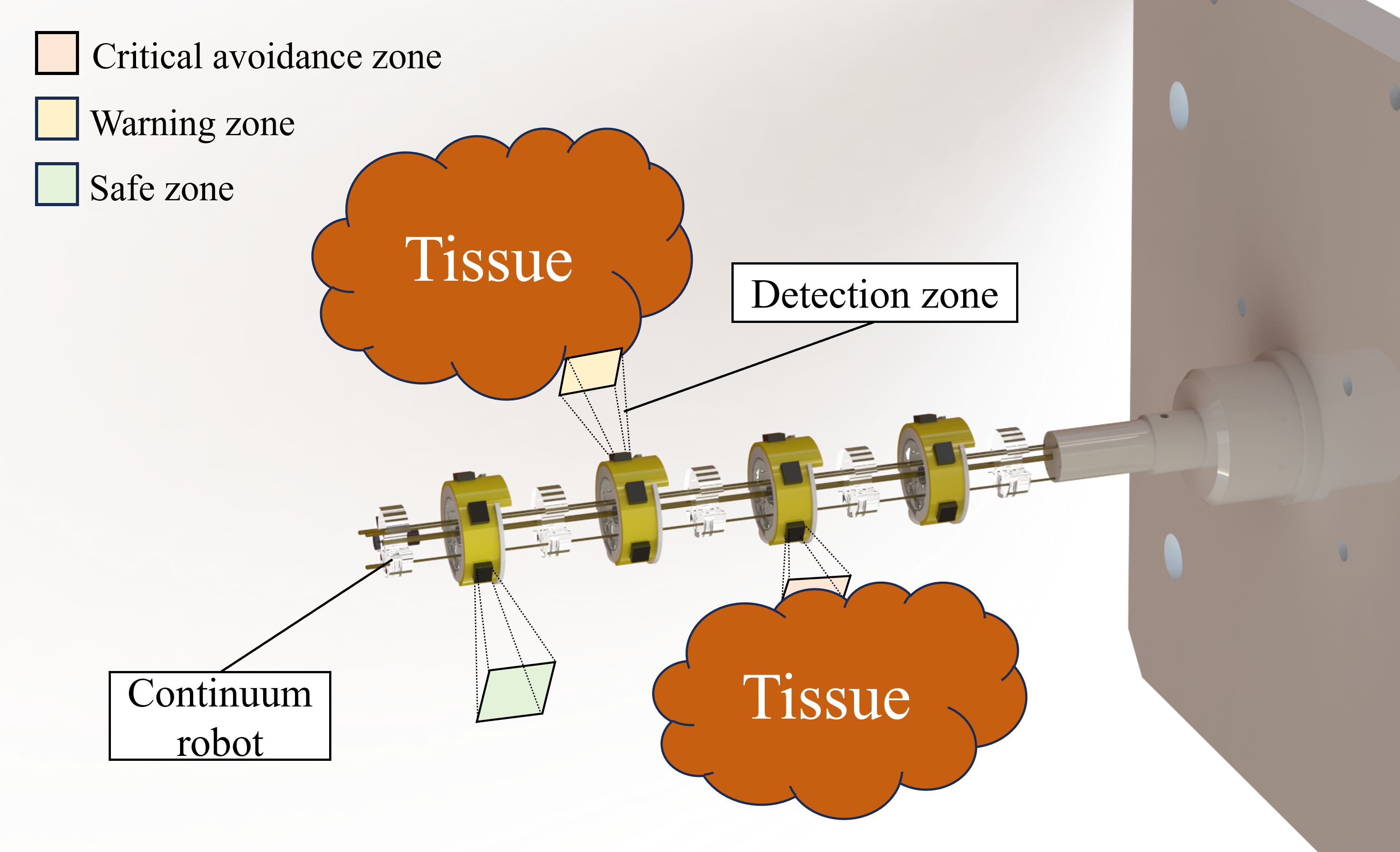}
    \caption{Schematic diagram of a continuum robot with integrated circumferential sensing.}
    \label{fig:RealEXP}
\end{figure}   

Despite these technological achievements, the comprehensive perception of peripheral anatomical obstacles remains unrealized. Although surgeons can visually assess forward facing tissues through endoscopic imaging, organ peristalsis-induced structural deformation can cause instrument embedding into soft tissues, resulting in iatrogenic injury. Therefore, in dynamic surgical settings, the ability to track robots in real time in a complex environment\cite{wang2015robot}, 3D reconstruction of surrounding anatomy\cite{cui2024towards}, and intelligent interaction strategies\cite{luo2024efficient} is of critical clinical relevance.

State-of-the-art flexible manipulators have integrated multimodal sensing technologies\cite{zhou2024integrated}, including magnetic sensors\cite{tmech2025fast}, optical shape sensors\cite{amirkhani2023design} and force/torque sensors\cite{donat2021real}. For example, Gao et al.\cite{gao2024body} proposed a tension profile sensing mechanism for collaborative multipoint estimation of contact forces and arm shapes in continuum manipulators. However, this approach suffers from temperature drift interference and response delays, which hinders real-time perception of peripheral environmental topology. Lv et al.\cite{lv2024fluoroscopy} implemented external force detection using distributed inductive curvature sensing combined with Euler-Bernoulli beam theory, yet failed to resolve three-dimensional spatial force vector decoupling. Donat et al.\cite{donat2021real} employed multiaxis force sensors for real-time concentric tube robot modeling, but their single external force point contact assumption limits applicability in multiorgan interaction scenarios. These collective limitations demonstrate that while existing methods enhance shape perception, they remain incapable of simultaneously resolving spatial orientation vectors and dynamic contact force amplitudes during obstacle collisions.

Whole-body perception for continuum robots has emerged as a significant research focus. Massari et al.\cite{massari2020machine} leveraged machine learning integrated with fiber Bragg grating (FBG) technology to create a soft tactile sensor capable of precise estimation of both magnitude and location of contact loads on compliant materials. Sun et al.\cite{sun2021triboelectric} introduced a self-powered triboelectric tactile sensing ring integrated onto continuum robot surfaces for external contact characterization. Jin et al.\cite{jin2021bioinspired} developed a liquid-metal-based triboelectric sensor for collision detection in bio-inspired soft robots. Complementing these, \cite{noh2019contact} implemented an optical intensity-based triaxial force sensor featuring S-shaped structures with modular sensing plates, while \cite{abah2021multi} achieved irregular spatial mapping through distributed sensing disk units deployed along robotic arms. However, these approaches face substantial limitations in medically constrained environments owing to bulky form factors and potential material hysteresis-induced drift phenomena.

Inspired by the optical intensity-based force sensor and micro force sensing system presented in \cite{noh2019contact} and \cite{yao2023rnn}, our work introduces an annular flexible sensor based on optical intensity sensing principles, integrated into an 8mm vertebral disk-actuated continuum robot. Each sensing unit independently detects local obstacles and reconstructs peripheral contours in real-time through coordinate transformation. Leveraging a radial scanning mechanism, our system quantitatively evaluates environmental features relative to robot-anchored sensors, enabling proactive collision prediction and obstacle avoidance. The primary contributions of this work include:
\begin{itemize}
    \item Modular flexible sensing architecture: We develop a flexible printed circuit board (FPCB) structure that equips each vertebral disk with standalone environmental perception capabilities, enhancing modular sensing in continuum robots.

    \item Real-time perception algorithm: An accompanying sensing algorithm is proposed and experimentally validated for environmental awareness, demonstrating measurement accuracy and operational robustness.

    \item Enhanced environmental awareness: Compared to previous work \cite{tmech2025fast}, our system improves the ability to perceive environmental obstacles of continuum robots in confined spaces. When integrated with existing navigation methods, it enables improved autonomous motion planning in complex luminal environments.
\end{itemize}

\section{System Model}
\begin{figure}[h]
    \centering
    \includegraphics[width=0.5\textwidth]{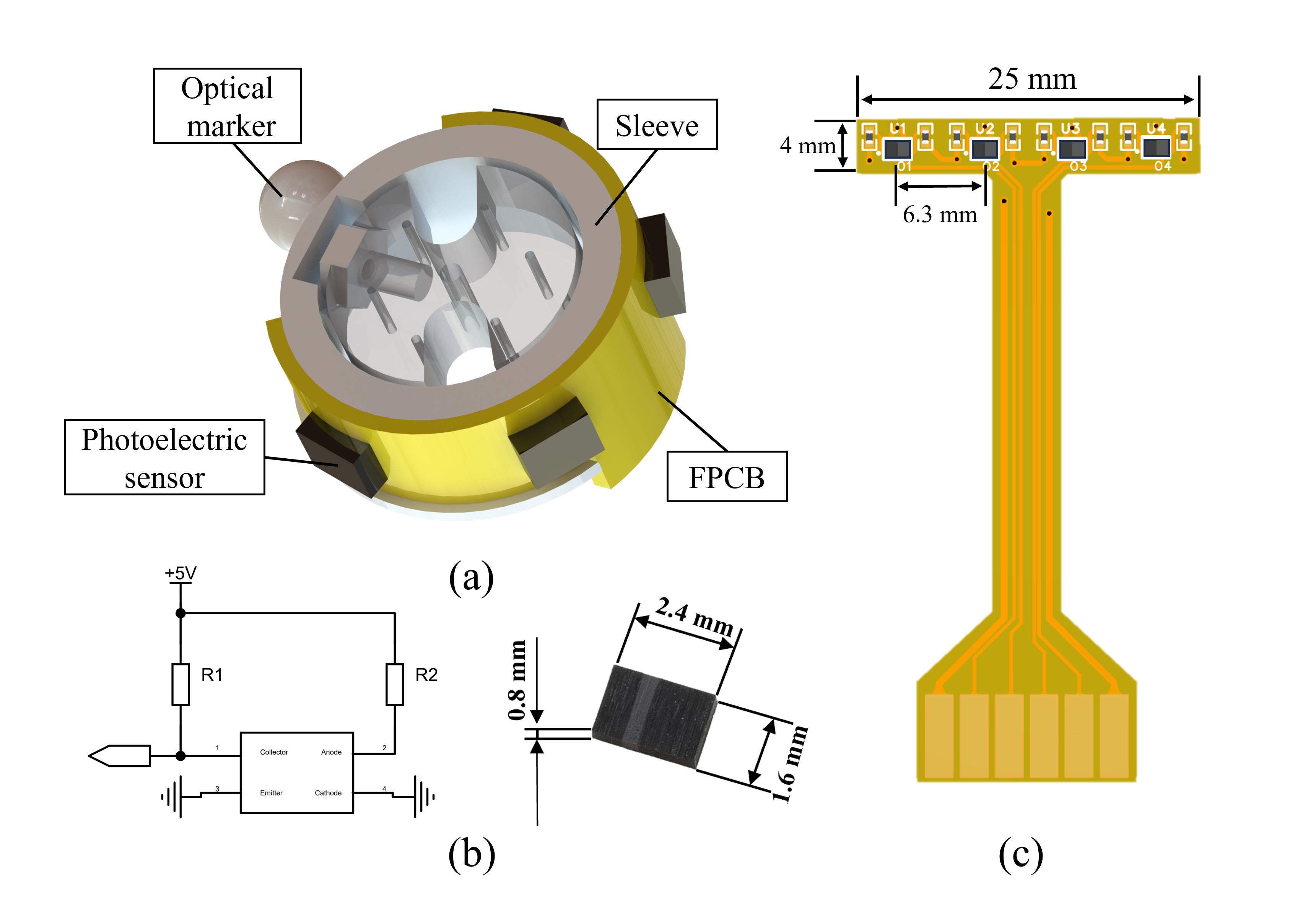}
    \caption{System model overview. (a) Schematic of the disk structure with integrated sensor modules. (b) Operational principle and dimensional layout of the sensing unit. (c) Appearance and configuration of the FPCB.}
    \label{fig:system_model}
\end{figure}

The sensor comprises two components (see Fig. \ref{fig:system_model}): a T-shaped flexible circuit board and NJL5901R-2 sensor units. The transverse section (25mm width $\times$ 4mm height) serves as the primary sensor deployment zone, while the longitudinal section integrates signal transmission and power supply lines. A 3D-printed sleeve mechanically couples the robot's vertebral disks with the sensor. The sensor concentrically wraps the sleeve, which features dedicated interfaces for managed wiring harness.

The system measures the key parameter $\Delta Z$, defined as the minimum Euclidean distance between the sensor and obstacles in the workspace. Photoelectric sensors emit constant-intensity light via LEDs, and phototransistors convert the reflected light intensity into voltage signals. Within anatomically constrained luminal environments, proximity of the luminal wall reduces the optical path length between the sensor and environment, consequently increasing the output voltage.

\begin{table}[htbp]  
  \centering  
  \caption{Cost of the Perimeter Sensing and Robotic System}  
  \label{tab:system_cost}  
  \begin{tabular}{l l l l}  
    \toprule  
    Structure & Type & Quantities & Cost \\  
    \midrule  
    Optoelectronic sensor & NJL5901R-2 & 4 & \$3 \\
    Flexible resistor  & 0402WGF1201TCE & 8 & \$1 \\
    FPCB & Self-designed & 1 & \$20 \\
    Supporting components & PLA 3D-printed & 50g & \$5 \\
    \multicolumn{3}{l}{Perimeter sensing system} & Total \$29 \\  
    \bottomrule  
  \end{tabular}
\end{table}

The four sensor units are angularly equidistant in their radial configuration upon installation. When external objects approach, the system detects quad-channel voltage variations, resolves environmental deformation via voltage-distance mapping, reconstructs contact orientation from distance data, and updates environmental maps in real time.

\section{Methodology}
\subsection{Sensor Calibration}
\begin{figure}[h]
    \centering
    \includegraphics[width=0.45\textwidth]{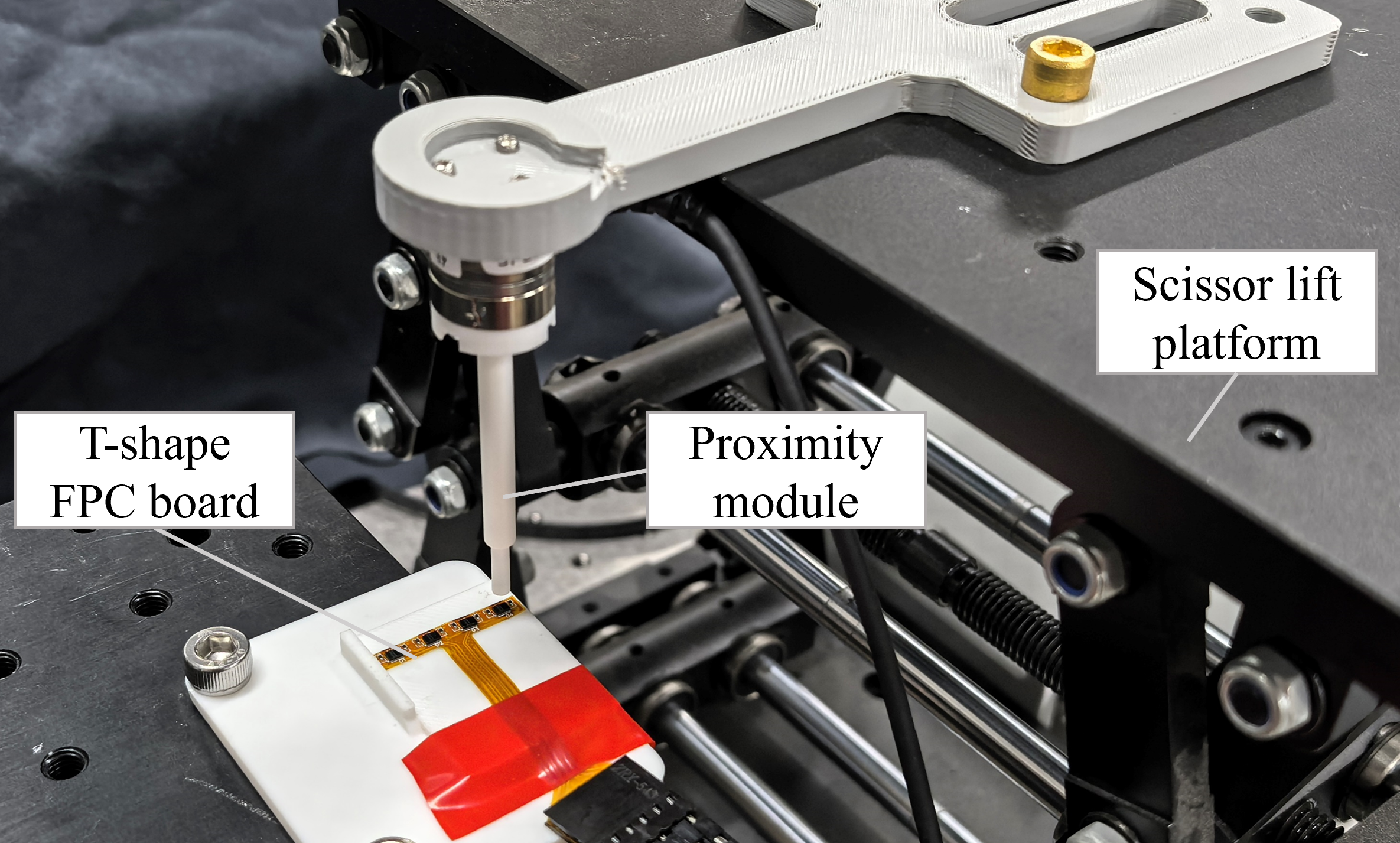}
    \caption{Sensor calibration platform with optical scissor-lift mechanism and precision displacement control.}
    \label{fig:test_platform}
\end{figure}

\begin{figure}[htbp]
    \centering
    \includegraphics[width=0.5\textwidth]{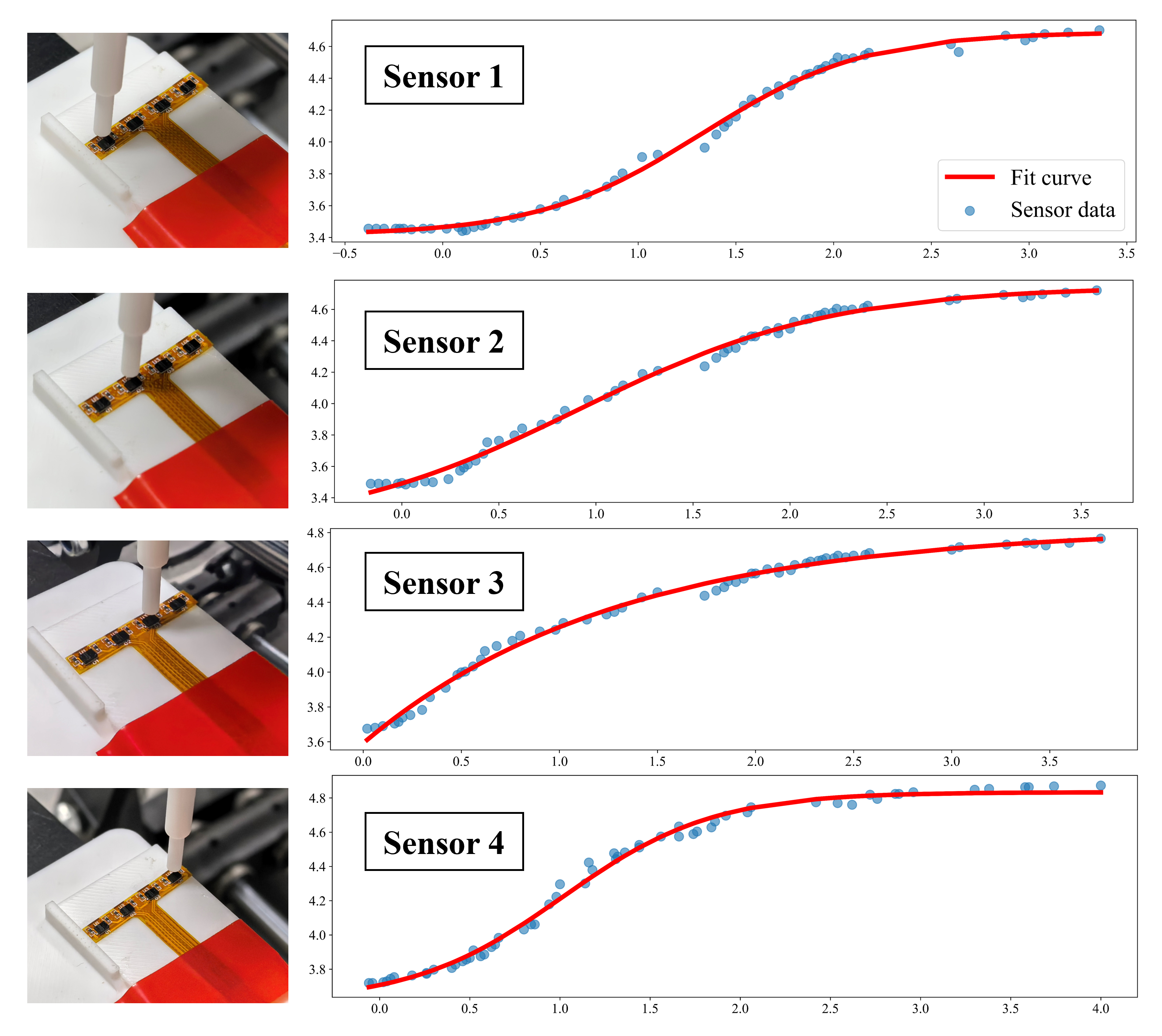}
    \caption{Calibration results and voltage-distance fitting curves for the four sensing units on T-shaped sensor.}
    \label{fig:logistic_fit}
\end{figure}

To establish the quantitative relationship between sensor output and environmental distance, this study employs a rapid calibration method based on an optical scissor-lift platform. As shown in Fig. \ref{fig:test_platform}, the calibration system fixes the proximity target to a precision displacement stage while anchoring the sensor to a base plate, with analog output acquired via an Arduino UNO board. The calibration process begins with verifying the stability of the measurements: 200 consecutive data samples were collected at 20 random positions. The variance of the raw data is approximately 1.38, and the variance of the sensor readings after moving-average noise reduction is 0.35. This confirms the excellent steady state performance of the sensor for acquiring data. Subsequently, distance-voltage characterization is performed, with 60 data points acquired per sensor unit (distribution shown as dots in Fig. \ref{fig:logistic_fit}).

\begin{equation}
V(\Delta Z) = \frac{V_{\mathrm{max}}}{1 + e^{-k(\Delta Z - \Delta Z_0)}} + \epsilon
\label{eq:logistic}
\end{equation}

Given the characteristic sigmoidal monotonic response of the sensor data, a logistic function is adopted for fitting, expressed in its canonical form by Eq. \ref{eq:logistic}. This model fully characterizes the response through three physical parameters: critical distance $\Delta Z_0$, saturation voltage $V_{max}$, and sensitivity coefficient k. Compared to polynomial and exponential fitting, this method exhibits superior parameter interpretability and noise immunity. The final fitting results demonstrated an average goodness-of-fit ($R^2$) exceeding 0.993 (response curve in Fig. \ref{fig:logistic_fit}).

\subsection{Discrete Circumferential Sensing for Continuum Robots}
\begin{figure}[ht]
    \centering
    \includegraphics[width=0.5\textwidth]{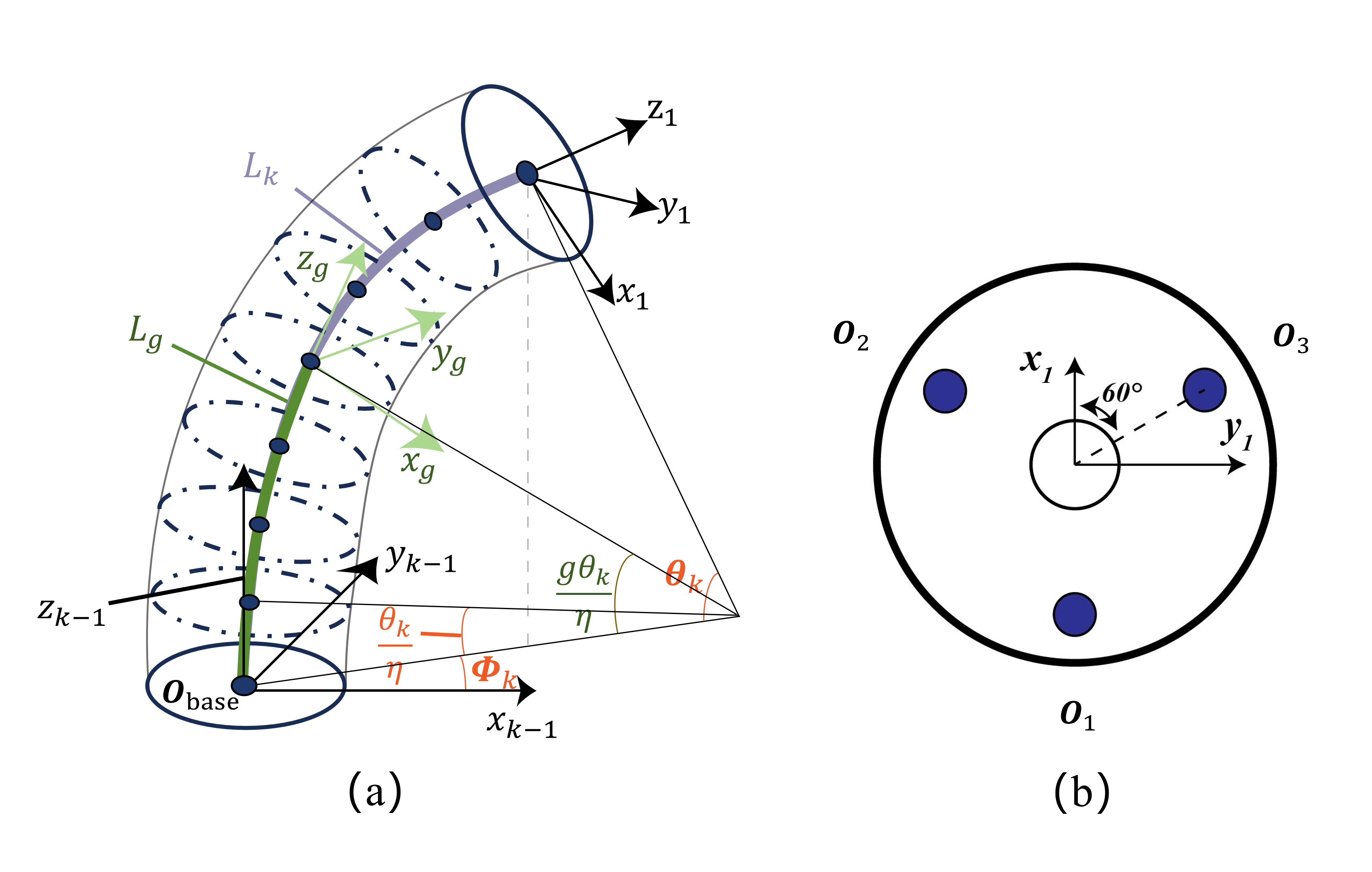}
    \caption{Kinematic modeling of a cable-driven continuum robot. (a) PCC model parameters and segment subdivision methodology. (b) Cable routing configuration and actuation scheme.}
    \label{fig:PCC}
\end{figure}

The sensor integration design requires explicit consideration of vertebral disk pose within the robot's coordinate frame. For the single-segment continuum robot used in this work, we employ the piecewise-constant curvature (PCC) model for precise geometric characterization. As shown in Fig. \ref{fig:PCC}, this modeling approach approximates the robot as a series of mutually tangent circular arcs with axially invariant curvature, where each segment's configuration is completely described by the parameter set ($\theta_k,\phi_k,R_k$):
\begin{equation}
    p_{k-1}^{k} = \frac{L_k}{\theta_k}
    \begin{bmatrix}
    (1-\cos{\theta_k}) \cos{\phi_k} \\
    (1-\cos{\theta_k}) \sin{\phi_k}  \\
    \sin{\theta_k}
    \end{bmatrix}
    \label{eq:eq1}
    \end{equation}

where the $\theta_k$ represents the bending angle, $\phi_k$ denotes the bending direction angle and $R_k$ serves as the bending radius of the $k^{th}$ segment. And $\theta_k \in [ 0, \pi ]$, $\phi_k \in [ -\pi, \pi ]$ and $R_k = L_k / \theta_k$. The end position of 
the $k^{th}$ segment can be formulated by these parameters as:
\begin{equation}
    \theta_k(\boldsymbol{q}_k) = \dfrac{2\sqrt{q_{k,1}^2+q_{k,2}^2+q_{k,3}^2-q_{k,1}q_{k,2}-q_{k,2}q_{k,3}-q_{k,1}q_{k,3}}}{3r}
    \label{eq:eq2}
\end{equation}

\begin{equation}
    \phi_k(\boldsymbol{q}_k) = \operatorname{atan2}(\sqrt{3}(q_{k,2}-q_{k,3}), q_{k,2}+q_{k,3}-2q_{k,1})
    \label{eq:eq3}
\end{equation}

Each robot segment requires at least 3 drive cables for full motion control. Precise adjustment of cable lengths enables controlled bending of the segment in any desired direction. The kinematic relationship between cable length variations and the robot's configuration parameters is given by:
\begin{equation}
    q_{k,i} = L_k - r\theta_k \cos{(\phi_k+(i-1)\xi)}
    \label{eq:eq4}
\end{equation}

Building upon the PCC model, the spatial pose distribution of disks can be computed by setting the subdivision parameter $\eta$ based on the number of disks (as depicted in Fig. \ref{fig:PCC}(a)). The pose parameters $C^{\boldsymbol{\zeta}_g}_{param}$ of the $g^{th}$($1 \leq g \leq \eta$) disk are obtained through the following equation:
\begin{equation}
C^{\boldsymbol{\zeta}_g}_{param} = \left[
\begin{array}{c}
x_g \\
y_g \\
z_g \\
\theta_g \\
\phi_g
\end{array}
\right]
= \left[
\begin{array}{c}
\frac{L_g}{\theta_g} (1-\cos{\theta_g}) \cos{\phi_g} \\
\frac{L_g}{\theta_g} (1-\cos{\theta_g}) \sin{\phi_g} \\
\frac{L_g}{\theta_g} \sin{\theta_g} \\
\frac{g}{\eta} \theta_k \\
\phi_k
\end{array}
\right]
\label{eq:eq_shape}
\end{equation}
where $L_g = \frac{g}{\eta}L_k$ and $\zeta_g$ denotes the pose of the $g^{th}$ disk.

\begin{figure}[h]
    \centering
    \includegraphics[width=0.48\textwidth]{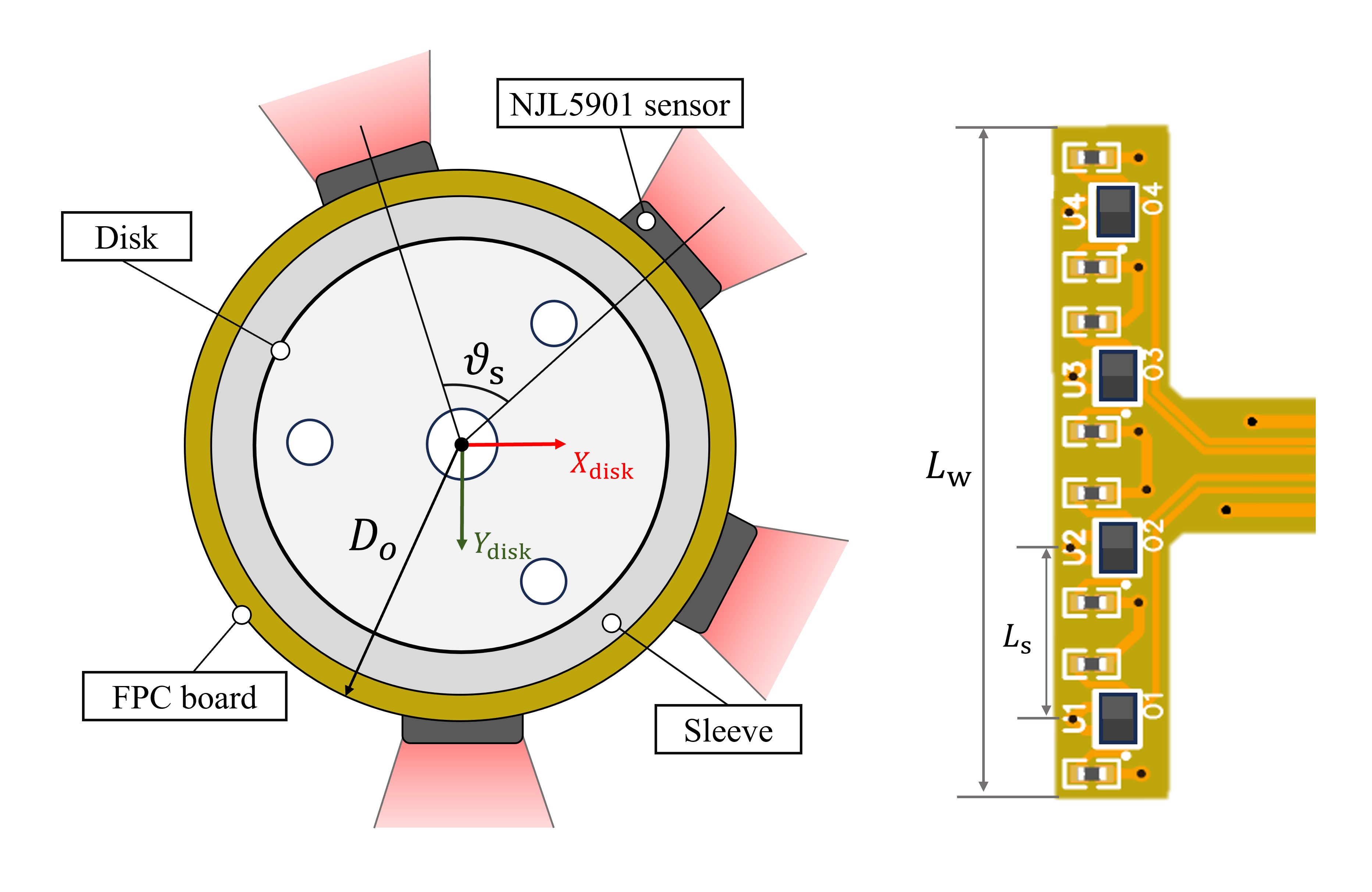}
    \caption{Top-view of sensor-disk integration and T-shaped sensor configuration.}
    \label{fig:env_reconstruction}
\end{figure}  

Building upon the resolved spatial poses of robotic disks, this study integrates sensors via sleeve structures to advance geometric calibration of sensing units. Establishing precise orientation relationships between each sensing element and the robotic coordinate frame is essential for real-time environmental reconstruction. As shown in the top-view installation diagram (See Fig. \ref{fig:env_reconstruction}), the transverse section of the T-shaped sensor is concentrically bonded to the sleeve via adhesive bonding, with four sensing units radially spaced at angular intervals of $\vartheta_s$:
\begin{equation}
    \vartheta_s = L_s / \pi D_o
    \label{eq:sensor}
\end{equation}
where $D_o$ denotes the sleeve outer diameter (in mm), and $L_s$ represents the center-to-center spacing between adjacent sensor units (in mm). This configuration achieves approximately 200 degree of azimuthal coverage. The detection range is intentionally limited due to lateral mounting of optical marker balls on the sleeve for ground-truth positional verification of vertebral disks. By synchronizing acquired vertebral pose (position/orientation), the spatial pose of each sensing unit in the world frame is computationally reconstructed, enabling environment mapping through multisensor fusion. Algorithm \ref{alg:alg1} summarizes the complete computational procedure. By continuously updating the local perception map, the updated map information $M_{update}$ can be integrated into the continuum robot navigation control pipeline, enabling safer robot-environment interaction.
\begin{algorithm}
  \LinesNumbered
  \caption{{Discrete Circumferential Obstacle Mapping  Algorithm.}}
  \label{alg:alg1}
  \KwIn{Raw sensor data $\mathcal{D}_{raw}$; Sensor parameters $C_{sensor}$; Robot actuation $q$; Subdivision number $\eta$; Sensor disk index $g$; Previous local map $M_{prev}$}
  \KwOut{Updated local obstacle map $M_{update}$}
  
  $\mathcal{D}_{filtered}$ $\leftarrow$ DataPreprocess($\mathcal{D}_{raw}$)\;
  
  Robot configurations $\Psi$ $\leftarrow$ Calculate robot kinematics $F(q)$\;

   Disk pose $P_{disk}$ $\leftarrow$ Calculate sensor disk pose $H(g, \eta,\Psi)$\;

   Sensor pose $p_{sensor}$ and orientation $o_{sensor}$ $\leftarrow$ Calculate sensor pose $G(C_{sensor},\Psi, P_{disk})$\;
  
 \For{each $reading$ in $\mathcal{D}_{filtered}$}{ \tcp{Map to calibrated distance $d_{gap}$}
    \If{$reading$ $<$ $threshold_{low}$}{
        $d_{gap}$ $\leftarrow$ $0$\; 
    }
    \ElseIf{$reading$ $>$ $threshold_{up}$}{
        $d_{gap}$ $\leftarrow$ MAX-DISTANCE\;
       } 
    \Else{
        $d_{gap}$ $\leftarrow$ CalibrationCurve($reading$, $C_{sensor}$)\;
        } 
        }
    
  $M_{update}$ $\leftarrow$ UpdateMap$(d_{map},p_{sensor}, o_{sensor}, M_{prev})$\;
  
  \Return $M_{update}$\;
Robot navigation module $\leftarrow$ $M_{update}$;
\end{algorithm}



\section{Experiment and discussion}
\begin{figure}[h]
    \centering
    \includegraphics[width=0.48\textwidth]{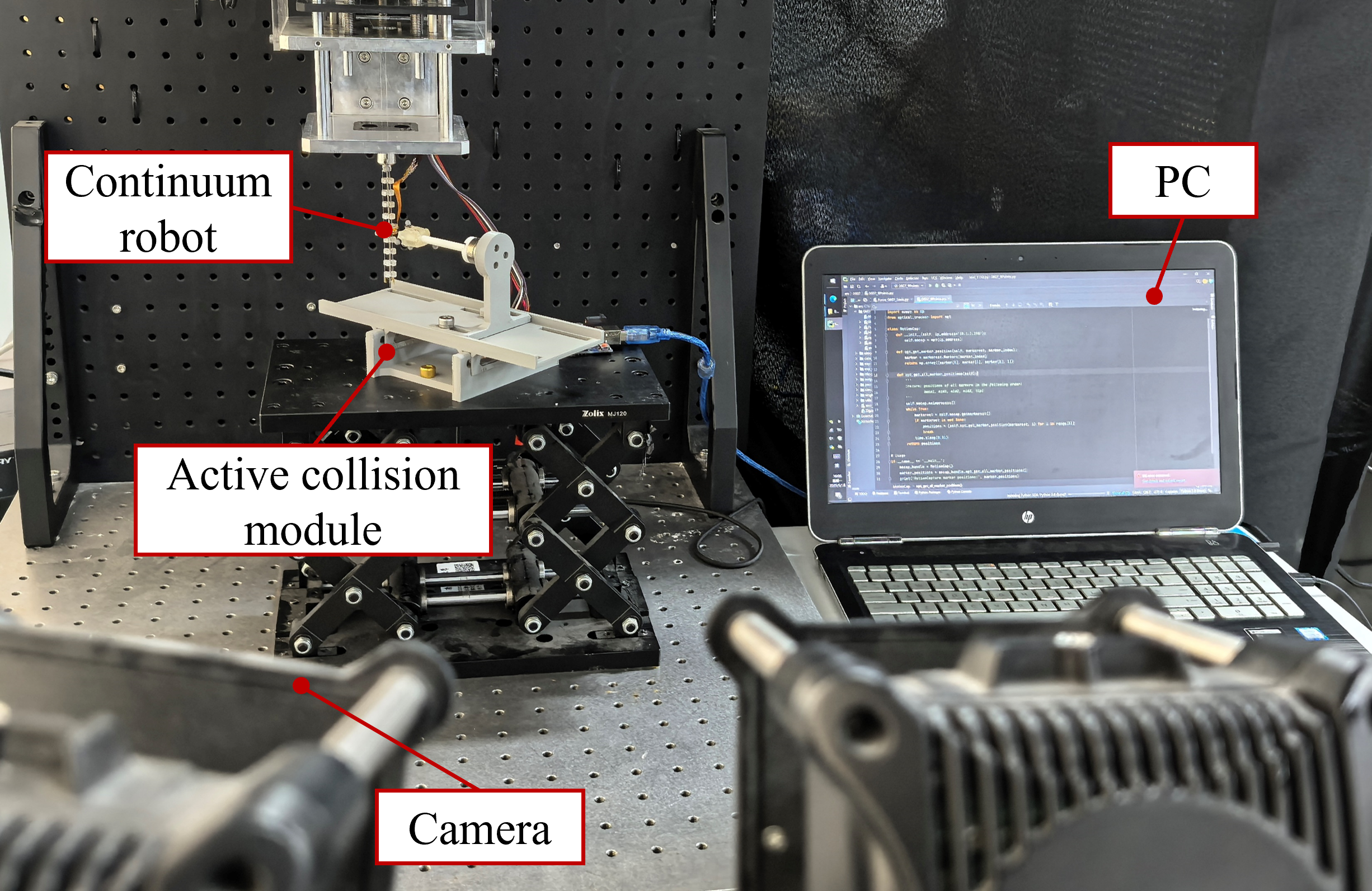}
    \caption{Comprehasive overview of the real-world experimental platform.}
    \label{fig:env_setup}
\end{figure}

In this section, we set up the operational platform in a physical environment (Fig. \ref{fig:env_setup}) and validated the proposed sensor system through a series of experiments. First, real-time data preprocessing was applied to acquired sensor signals during measurements. Second, structurally simulated obstacles were maneuvered in proximity to the continuum robot, allowing comparative analysis between sensor output and ground-truth optical measurements to verify the reliability of environmental perception. Finally, after integration with the robotic control system, the sensor enables real-time obstacle detection in the robot's vicinity while simultaneously updating the environment map with sensory feedback. These updated mapping data facilitate autonomous obstacle avoidance, thereby achieving efficient navigation in confined workspaces.

\subsection{Sensor Signal Preprocessing}
\begin{figure}[h]
    \centering
    \includegraphics[width=0.49\textwidth]{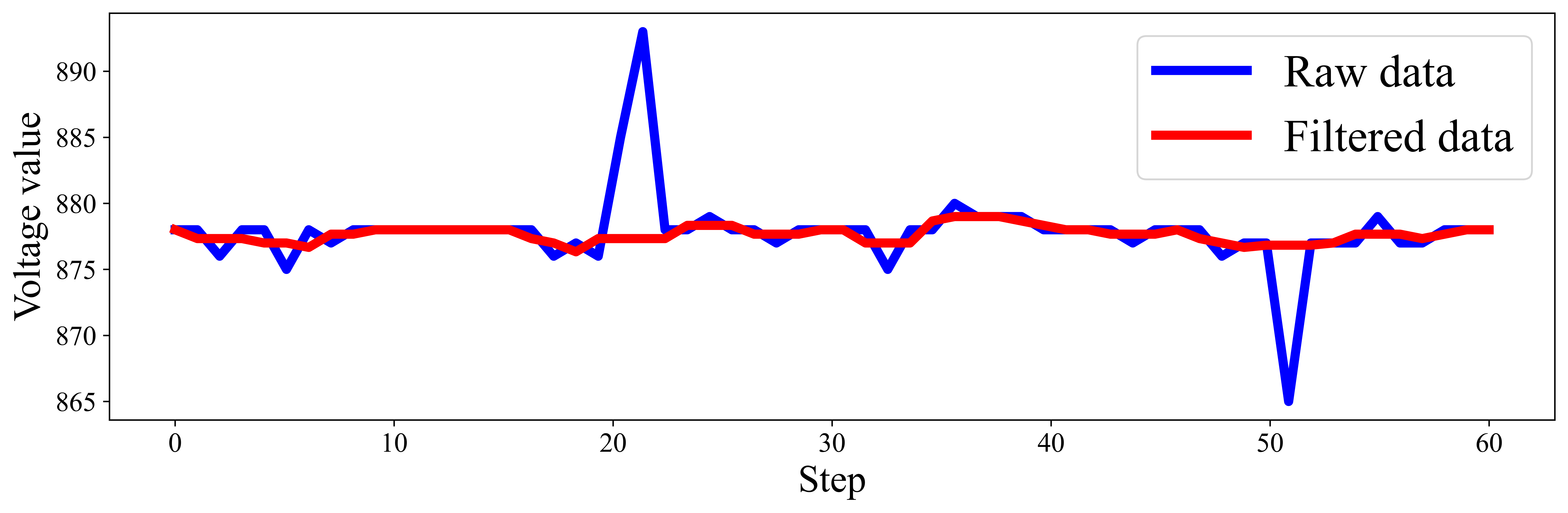}
    \caption{Comparison of raw and filtered sensor data (representative data).}
    \label{fig:real_exp1}
\end{figure}

Within the previously calibrated measurement range, the sensor demonstrated stable distance detection capability at fixed positions. Therefore, a hybrid filtering scheme that combined moving average and interquartile range (IQR)-based outlier detection was used for data acquisition. The filtering effect of one set of data is shown in Fig.\ref{fig:real_exp1}. Following sampling of the raw voltage signal at 100Hz, the measurement data was re-disoised in real time using a moving average filter with window size $N=10$. This was followed by the rejection of statistical outliers based on the interquartile range (IQR). Data points exceeding the IQR threshold were deemed invalid to obtain stable sensor readings. Compared to original data, the filter signal quality index (SQI) improved by 13\%, producing higher quality data for subsequent applications.

\subsection{Perception Reliability Validation}
\begin{figure}[h]
    \centering
    \includegraphics[width=0.5\textwidth]{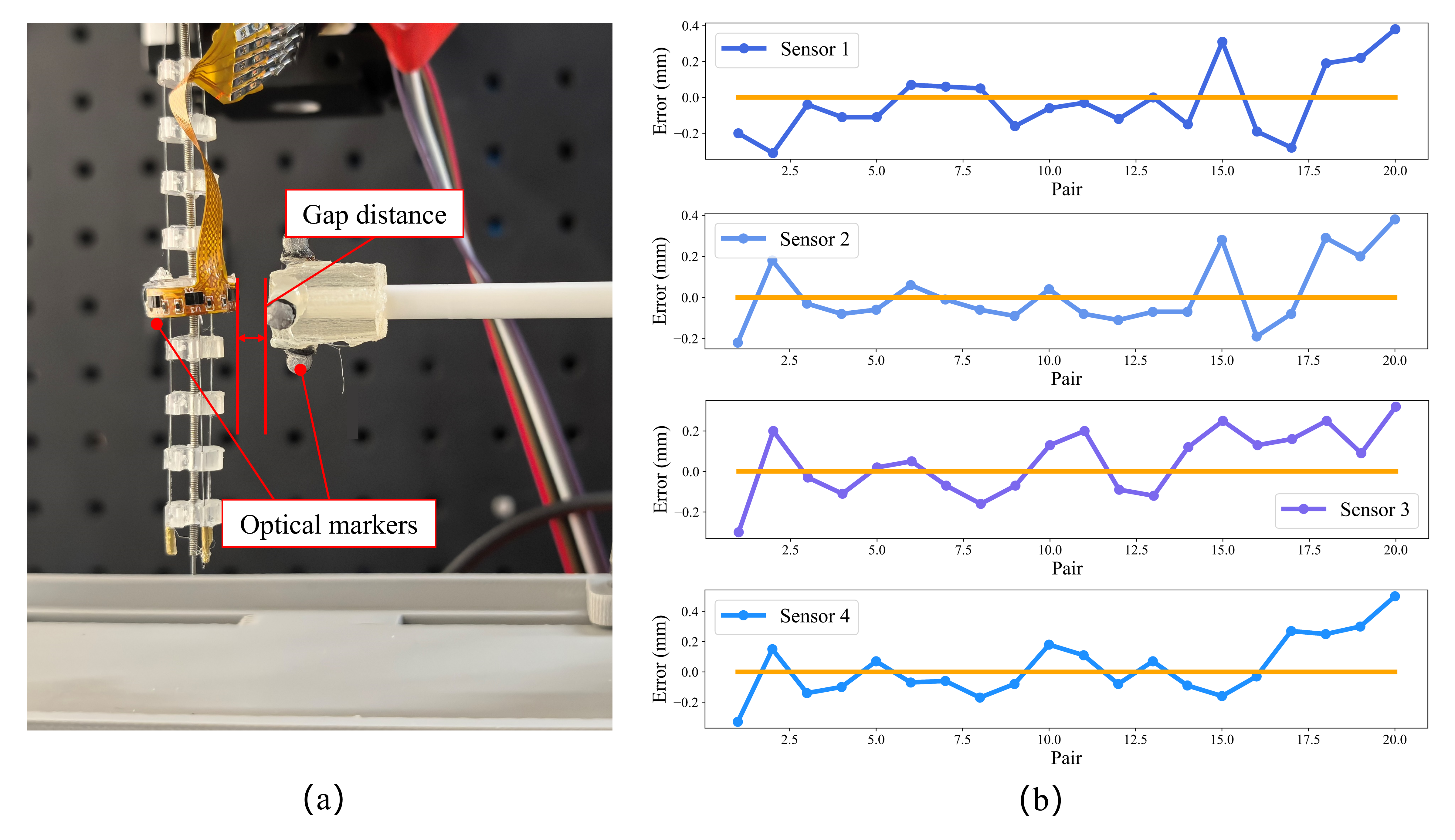}
    \caption{Perception reliability experiment: (a) Process of an obstacle approaching the sensor. (b) Localization error values when approaching four sensors in different test groups}
    \label{fig:real_exp2}
\end{figure}

By correlating the sensor output with the calibration curves, the estimated distance between obstacles and the sensing units is derived. As shown in the experimental platform in Fig. \ref{fig:real_exp2}(a), a device performs controlled proximity maneuvers toward the robotic disk, eliciting characteristic real-time response signatures. Synchronized pose data acquisition via the external optical motion capture system (Mars 4H, $\text{NOKOV}^\circledR$, China) establishes precise ground-truth distance references. This system operates at a sampling rate of 180 Hz and achieves a 3-D accuracy of $\pm0.1$mm. We control the obstacle’s approach toward the sensor using a proximity module, measuring the obstacle distance in real time via sensor readings and recording its error relative to the ground truth from an optical camera. The selected proximity module meets the requirements for optical reflection measurements, making it suitable for reflection-based sensor detection. Fig. \ref{fig:real_exp2}(b) quantitatively demonstrates the ranging error distribution across four sensing units during multidirectional approach scenarios, where the postcalibration root mean square error (RMSE) reaches 0.20mm, validating the reliability of the system.

\subsection{Real-time Map Updating}
\begin{figure}[h]
    \centering
    \includegraphics[width=0.5\textwidth]{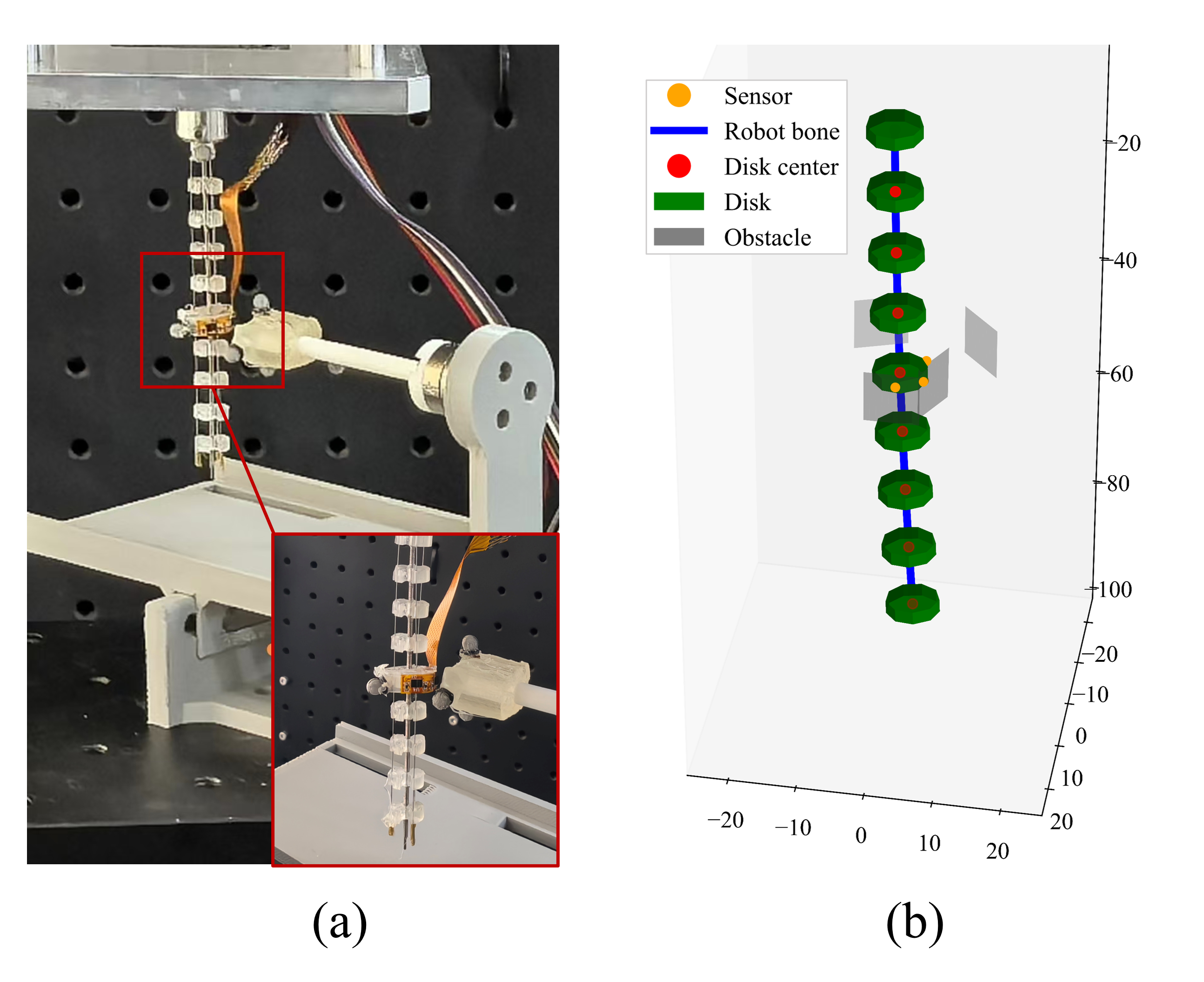}
    \caption{Map update process: (a) Experimental demonstration of dynamic obstacle approach toward the robotic platform. (b) Real-time mapping results showing reconstructed obstacle planes from sensor data.}
    \label{fig:real_exp3}
\end{figure}

Integrated with the continuum robot's control system, the proposed perception system updates environmental maps in real-time when obstacles approach preset thresholds. As illustrated in Fig. \ref{fig:real_exp3}(b), the sensor feedback constructs multiple planes at varying distances, which represent the closest obstacle positions relative to the robot. By dynamically annotating circumferential safety zones, warning zones, and obstacle intrusion zones with different clearance thresholds, this approach establishes a fundamental environmental interaction model for collision avoidance decision-making. The complete pipeline latency for map updates is limited to less than 200 ms---validated through collision simulation experiments where the system responded to intrusion events within 185$\pm$22 ms when contact modules simulated obstacle impacts. The execution of obstacle avoidance follows the motion planning framework in \cite{luo2024efficient}, while this work focuses on validating the real-time capabilities and reliability of the map generation module. These experiments ultimately confirm that the proposed sensing system enables dynamic circumferential environment monitoring, establishing the technical foundation for high-precision control of continuum robots in complex operational scenarios.

Although the results of this work demonstrate encouraging potential, certain aspects require further refinement to advance this research direction. First, the current experiments only optimize a single sensing disk along the robot's deployment. Future work will involve multi-disk collaborative calibration to further improve environmental perception accuracy. Second, the existing architecture lacks quantitative force sensing capability. Subsequent improvements will integrate flexible piezo resistive sensing structures to achieve dual-mode distance force detection. Finally, proprioception of the robot and environmental awareness have not yet been integrated. Future work plans to fuse circumferential obstacle perception with continuum robot shape sensing based on permanent magnetic sensing technology \cite{tmech2025fast} and endoscopic vision into a body-peripheral-end framework for reliable navigation in complex anatomy. This will lay the foundation for active obstacle avoidance in complex anatomical environments.

\section{Conclusion}
In this work, we present a novel sensor capable of independently perceiving its surrounding environment for real-time detection when integrated into continuum robots. The sensor integrates four sets of photoelectric sensor units using flexible printed circuit board technology. By acquiring sensor readings in real time, it achieves real-time detection of radial distances along the robot's circumference. Experiments validated the feasibility and accuracy of the proposed method for environmental perception upon integration with the robot. This modular distance detection scheme can be applied to more scenarios where continuum robots are used once, and its lightweight and low cost characteristics can provide more options for medical applications. The results demonstrate errors between the actual distance approaching and the sensor-reconstructed distance, confirming the effectiveness of the sensor design for real-time environmental perception. Future developments will focus on sensor miniaturization and contact-force sensing structures to provide precise haptic feedback during collisions in anatomically constrained spaces such as the intestinal tract, enabling accurate confined-space control.

\bibliographystyle{reference/IEEE}
\bibliography{reference/ref.bib}

\end{document}